# Silicon Sampling via Cross-Survey Transfer

Chan-Tung Ku
Department of Information Management
National Sun Yat-sen University
Kaohsiung, Taiwan
kuchantung@gmail.com

Chan Hsu
Department of Information Management
National Sun Yat-sen University
Kaohsiung, Taiwan
chanshsu@gmail.com

Pei-Cing Huang
Department of Information Management
National Sun Yat-sen University
Kaohsiung, Taiwan
pcpeicing@gmail.com

Frank Cheng-shan Liu
Institute of Political Science
National Sun Yat-sen University
Kaohsiung, Taiwan
csliu@mail.nsysu.edu.tw

I-Ling Cheng
Graduate Institute of Library and Information Science
National Chung Hsing University
Taichung, Taiwan
chengi428@gmail.com

Yihuang Kang
Department of Information Management
National Sun Yat-sen University
Kaohsiung, Taiwan
ykang@mis.nsysu.edu.tw

***Abstract*—Silicon sampling—using large language models (LLMs) to simulate human survey respondents—has emerged as a promising approach for augmenting traditional survey research. However, most evaluations rely on distributional comparisons rather than individual-level prediction, which risks conflating pattern matching with coherent respondent-level prediction. We propose cross-survey transfer, a more rigorous evaluation framework in which an LLM is given a respondent's answers to one set of questions and must predict their answers to entirely different questions from the same survey. Using data from the Taiwan Election and Democratization Study (TEDS) 2024, three open-weight LLMs (27B–120B parameters), and supervised machine learning baselines, we find that: (1) zero-shot LLMs achieve 52% accuracy on genuinely unseen items, closing to within 6 percentage points (pp) of a supervised random forest trained on same-population data; (2) a stable construct predictability hierarchy emerges, from 67% for partisan attitudes to 23% for sovereignty; and (3) variance collapse and safety alignment effects—two commonly cited LLM limitations—turn out to be more nuanced than previously reported, with variance collapse affecting supervised models as well and alignment effects varying dramatically across model families. These findings clarify both the promise and boundaries of silicon sampling.**

## I. Introduction

Survey research faces growing practical challenges. Response rates have declined substantially over recent decades, while costs per completed interview continue to rise [1]. At the same time, policymakers and researchers increasingly need rapid, fine-grained measurements of public opinion—precisely when traditional methods are becoming slower and more expensive.

In this context, the idea of using large language models (LLMs) as simulated survey respondents—termed *silicon sampling* by Argyle et al. [2]—has attracted considerable attention. The premise is straightforward: condition an LLM on a demographic profile and attitudinal context, then ask it to complete a survey as that persona. If the simulated responses approximate real ones, silicon sampling could dramatically reduce the cost and turnaround time of survey research. Early results have been encouraging, with LLMs reproducing aggregate opinion distributions from real surveys [2] and predicting experimental treatment effects with striking accuracy [3].

However, a methodological concern undermines much of this optimism. Most evaluations condition LLMs on demographic profiles and assess performance at the distributional level—comparing aggregate response distributions between silicon and human samples [2], [4], [5]. Even the most rigorous designs, such as Argyle et al.'s Study 3 (which uses 11 survey items to predict a 12th), evaluate distributional correspondence rather than individual-level prediction accuracy. This means we know LLMs can reproduce population-level patterns, but not whether they can support coherent respondent-level prediction across different constructs—a harder and more meaningful test.

This paper asks a harder question: *Can LLMs predict an individual respondent's attitudes on topics they have never seen, given only that respondent's answers to a different set of questions?* We formalize this as cross-survey transfer—partitioning survey items into disjoint context (Set A) and prediction (Set B) sets—and argue it provides a more rigorous, individual-level test of respondent-level generalization.

We evaluate this framework on the Taiwan Election and Democratization Study (TEDS) 2024, a nationally representative survey covering political engagement, institutional trust, democratic values, and cross-strait relations. Our choice of a Mandarin-language, East Asian democracy is deliberate: existing evaluations overwhelmingly focus on English-language surveys from Western, Educated, Industrialized, Rich, and Democratic (WEIRD) populations [5]–[7], leaving open the question of whether silicon sampling generalizes to other cultural and linguistic contexts. Taiwanese political surveys also involve constructs, such as sovereignty preferences on the independence–unification spectrum, which have no direct Western analogue, providing a stringent test of cross-cultural transfer.

To contextualize LLM performance, we compare three open-weight models against supervised machine learning baselines trained on the same features, and conduct abliteration (alignment-removal) experiments to disentangle the effects of safety training from base model capabilities. Our study is guided by three research questions:

*RQ1.* How accurately can zero-shot LLMs predict survey responses on genuinely unseen items, and how does this compare to supervised baselines with access to same-population training data?

*RQ2.* What constructs are more or less amenable to cross-survey transfer, and does a stable predictability hierarchy emerge across models?

*RQ3.* Are two commonly cited limitations of silicon sampling—variance collapse and safety alignment distortion—inherent to LLMs, or do they reflect broader properties of the prediction task and vary across model families?

The contributions of this paper are: (1) a novel individual-level evaluation framework based on cross-survey transfer; (2) disjoint construct partitioning as a test of out-of-construct generalization; (3) evaluation in a non-WEIRD, Mandarin-language political context; and (4) empirical analysis of variance collapse and alignment effects across model families.

The rest of this paper is organized as follows. Section II reviews the silicon sampling literature and identifies the evaluation gap our framework addresses. Section III presents the cross-survey transfer framework. Section IV describes our experimental setup, and Section V reports the results. We discuss implications and limitations in Sections VI and VII, and conclude in Section VIII.

## II. Background and Related Work

The idea of using LLMs as survey respondents originated with Argyle et al. [2], who coined the term "silicon sampling" and demonstrated that GPT-3, prompted with demographic backstories, could reproduce opinion distributions from the American National Election Study (ANES). Their evaluation assessed distributional correspondence—comparing aggregate patterns via Cramer's V—rather than individual-level prediction accuracy, and explicitly noted they "do not evaluate correspondence at the individual level." In their most stringent test (Study 3), 11 survey items served as interview-style context to predict a 12th, but evaluation remained distributional. Since then, the field has expanded along two axes. On the experimental side, Hewitt et al. [3] showed that GPT-4 simulations of treatment effects across 70 pre-registered experiments correlated strongly with actual effects ($r_{\text{adj}} = 0.91$), though this measures aggregate effect-size prediction, not individual response accuracy. On the individual side, Park et al. [8] achieved 85% normalized accuracy (relative to human test-retest reliability) on the General Social Survey by conditioning on 2-hour qualitative interviews—a rich but expensive input modality. Aher et al. [9] replicated classic social science experiments with LLM participants, again evaluating at the aggregate level.

Silicon sampling is a specific instance of the broader field of LLM social simulation, which uses language models to simulate human behavior in social contexts. Anthis et al. [10] recently identified five tractable challenges for this field: diversity, bias, sycophancy, alienness, and out-of-distribution (OOD) generalization. They argued that OOD generalization—the ability to simulate human responses in settings not well-represented in training data—is the most fundamental barrier. Our cross-survey transfer framework directly operationalizes this challenge: by requiring LLMs to predict responses on items they have never seen, we test precisely whether they can generalize beyond their training distributions.

Alongside these advances, several systematic limitations have emerged. Bisbee et al. [4] showed that LLM-generated response distributions are systematically narrower than real human distributions—a phenomenon we term variance collapse. Santurkar et al. [5] documented that LLM opinions cluster around younger, liberal, WEIRD demographics, and Cheng et al. [11] found that simulations of minority groups tend to stereotypically "caricature." Durmus et al. [12] measured representational gaps across 36 countries. On the alignment front, Perez et al. [13] showed that Reinforcement Learning from Human Feedback (RLHF) introduces systematic behavioral patterns, and work on preference collapse [14] suggests safety alignment may narrow output distributions. However, no prior study has systematically compared aligned and abliterated variants of the same models on the same survey task.

Recent work has begun addressing some of these limitations through fine-tuning [6], [15] and cross-cultural evaluation [7], but these studies largely retain distributional evaluation paradigms. Our work differs from prior approaches in three ways. First, we move beyond distributional evaluation to individual-level cross-construct prediction, providing a genuinely out-of-distribution test without requiring expensive interview data. Second, we include supervised ML baselines to contextualize LLM performance—a comparison absent from most silicon sampling studies. Third, we conduct abliteration experiments across multiple model families, enabling controlled comparison of alignment effects. Together, these design choices address the evaluation gap identified by Anthis et al. [10] while grounding our analysis in a non-WEIRD, non-English political context.

## III. Cross-Survey transferFramework

The goal of this paper is to evaluate whether LLMs can support coherent respondent-level prediction across constructs—not merely exploit local item correlations. To this end, we propose cross-survey transfer, a dataset-agnostic evaluation framework that tests LLM survey simulation under genuinely out-of-distribution conditions.

The framework begins with a partition of survey items. Let $S = \{q_1, \dots, q_n\}$ be a survey instrument and $D = \{d_1, \dots, d_m\}$ a set of demographic variables. We partition items into disjoint sets: Set A (persona-building context shown to the model) and Set B (prediction targets never shown). Each Set B item has at least one anchor in Set A measuring a related construct facet. The partition is theory-driven: we assign items to ensure each construct appears in both sets, with Set A containing items that, following attitude constraint theory [16], are expected to predict

TABLE I. SET B PREDICTION TARGETS: QUESTION TEXT, RESPONSE OPTIONS, AND SET A ANCHORS (RELATED ITEMS PROVIDED AS PERSONA CONTEXT; CODES FOLLOWS TEDS QUESTIONAIRE NUMBERING).

| ID | Question | Responses | Construct | Anchors |
|---|---|---|---|---|
| B3 | How much do you care about the presidential election? | 1=Very much, 2=Somewhat, 3=Not much, 4=Not at all | Political participation | B1, B2a |
| C2 | Did the Tsai government handle COVID-19 well? | 1=Very well, 2=Fairly well, 3=Not very well, 4=Not well at all | Government performance | C1 |
| C4 | Satisfied with the healthcare system? | 1=Very satisfied, 2=Fairly satisfied, 3=Not very satisfied, 4=Not at all satisfied | Political system | C3 |
| D2 | Officials don't care what people like me think. | 1=Strongly agree, 2=Agree, 3=Disagree, 4=Strongly disagree | Political efficacy | D1. D3 |
| D5 | Government wastes a lot of tax money. | 1=Strongly agree, 2=Agree, 3=Disagree, 4=Strongly disagree | Government trust | D4, D6 |
| D11 b | Trust the executive branch? | 1=Very much, 2=Somewhat, 3=Not much, 4=Not at all | Institutional trust | D11a,c |
| D11e | Trust political parties? | 1=Very much, 2=Somewhat, 3=Not much, 4=Not at all | Institutional trust | D11a,c |
| D15 | Having a strong leader who does not have to bother with parliament or elections would be … | 1=Very good, 2=Fairly good, 3=Neither good nor bad, 4=Fairly bad, 5=Very bad | Democratic attitudes | D13 |
| D20 | Which party best improves welfare? | DPP / KMT / TPP / None / Other | Party competence | D18, D19 |
| F2 | Economic outlook for next year? | 1=Much better, 2=Somewhat better, 3=About the same, 4=Somewhat worse, 5=Much worse | Economic evaluation | F1 |
| J2b | Like or dislike Hou Yu-ih? | 0=Strongly dislike to 10=Strongly like | Candidate evaluation | J2a, J2c |
| M2 | Cross-strait relations trend? | Better / About the same / Worse | Cross-strait relations | M1, M3 |
| M5a | Independence–unification position? | 0=Independence to 10=Unification | Sovereignty | N1a,4a,5a |
| P2b | Like or dislike the DPP? | 0=Strongly dislike to 10=Strongly like | Party favorability | P2a, P2c |

Set B responses through shared ideological structure. We assign approximately two-thirds of items per construct to Set A and one-third to Set B, and the partition is fixed before any experiments—not optimized post hoc.

Given this partition, we construct a prompt for each respondent containing four components: (1) a demographic profile in natural language; (2) the respondent's verbatim Set A answers with full question text and labels; (3) the Set B questions with response options; and (4) population-level response distributions for base-rate calibration, providing the model with realistic priors without any individual-level leakage.

To appreciate why this framework is a harder test than existing approaches, it helps to consider how prior evaluations work. Most silicon sampling studies condition the LLM on a respondent's demographic backstory and assess performance by comparing aggregate response distributions [2], [4], [5]. Even Argyle et al.'s most rigorous test evaluated distributional correspondence rather than individual-level accuracy. Park et al. [8] did evaluate at the individual level, but required 2-hour qualitative interviews as input—a costly modality that does not scale. In our supervised ML setting, models train on labeled examples from the same population via k-fold cross-validation, a strong baseline that requires in-domain data. Our proposed cross-survey transfer differs from both: the LLM operates entirely zero-shot, conditions on attitudinal context (not just demographics), and is evaluated on individual-level accuracy across different constructs—analogous to a pure domain-transfer problem.

We evaluate predictions at three levels: item-level (exact match, within-±1, MAE), construct-level (accuracy by theoretical construct), and distribution-level (variance ratio VR $= \sigma_{pred}/\sigma_{true}$, where values below 1.0 indicate collapse). All metrics are compared against a majority-vote baseline.

## IV. EXPERIMENTS

We apply the framework to the Taiwan Election and Democratization Study (TEDS) 2024 [17], a nationally representative telephone survey (*N = 1,206*) conducted after the January 2024 presidential and legislative elections. TEDS covers political engagement, government evaluation, institutional trust, democratic values, economic perceptions, cross-strait relations, and partisan attitudes.

We use 5 demographic variables, 28 Set A items (context), and 14 Set B items (prediction targets) spanning 13 constructs. Table I lists the Set B items with their question text, response options, constructs, and Set A anchors. Anchors are Set A items measuring a related facet of the same construct; item codes follow TEDS questionnaire numbering. For example, B3 (election concern) is anchored to B1 (voter turnout) and B2a (campaign attention), while M5a (sovereignty position) is anchored to N1a (national identity), N4a (unification pace), and N5a (independence pace)—all available to the model as persona context. After excluding respondents with excessive missing data, between 593 and 610 valid respondents were retained per prediction target, depending on item-specific missingness.

Several items involve constructs unique to Taiwanese politics. M5a places respondents on a 0–10 independence–unification scale—the central political cleavage in Taiwan, with no direct U.S. or European parallel. D20 asks which of three parties (DPP, KMT, TPP) best improves welfare, reflecting a multi-party system. J2b measures candidate favorability for Hou Yu-ih (KMT), requiring inference from party affect.

TABLE II. CROSS-SURVEY TRANSFER ACCURACY. ML BASELINES USE 5-FOLD CV (SUPERVISED); LLMS ARE ZERO-SHOT. VR = VARIANCE RATIO.

| | Method | Exact | ±1 | MAE | VR |
|---|---|---|---|---|---|
| ML | Majority vote | 46.6% | – | – | – |
| | LR (demo only) | 47.0% | 83.1% | 1.35 | 0.38 |
| | RF (demo+SetA) | **58.3%** | **88.2%** | **0.95** | 0.72 |
| LLM | Qwen3.5 27B | 52.1% | 85.7% | 1.17 | 0.84 |
| | gpt-oss 120B | 51.3% | 85.2% | 1.16 | **0.85** |
| | Gemma3 27B | 47.6% | 86.2% | 1.29 | 0.67 |

We evaluate three open-weight LLMs chosen for architectural diversity, multilingual coverage, and the ability to perform abliteration experiments (impossible with closed-source models):

- gpt-oss (120B, Q4) [20]: a reasoning model with chain-of-thought inference, providing a large-scale anchor.
- Qwen3.5 (27B, Q4) [21]: multilingual model with strong Chinese capabilities, testing whether language-specific training improves performance on Mandarin surveys.
- Gemma3 (27B, FP16) [22]: Google's model, providing an English-centric comparison point.

All models run locally via Ollama [18] on two NVIDIA DGX Spark GB10 systems (128GB unified memory). For each model, we also test an abliterated variant with safety alignment removed via directional ablation of the refusa direction [19], yielding six LLM conditions total.

To consider LLM performance, we include two supervised ML baselines trained on the same input features: (a) Logistic Regression (LR) on demographics only, and (b) random forest (RF; 200 trees, max depth 10) on demographics + Set A. These baselines use 5-fold cross-validation on the full *N = 1,206* sample, meaning each model trains on~960 labeled examples from the same population before predicting the held-out fold. This creates a fundamental information asymmetry: ML baselines perform in-domain supervised prediction, while LLMs operate entirely zero-shot. The ML baselines thus represent an upper bound—the accuracy achievable when the same-population labeled data is available.

## V. RESULTS

### A. Overall Performance (RQ1)

Table II presents all conditions. Random forest (RF) achieves the highest accuracy at 58.3%. Among zero-shot LLMs, Qwen3.5 leads at 52.1%, followed by gpt-oss at 51.3%. Logistic regression on demographics alone (47.0%) barely exceeds the majority baseline (46.6%), confirming that demographics provide minimal predictive signal—the information is in Set A responses.

Fig. 1 visualizes the accuracy comparison across information regimes. The 6 pp gap between RF and the best LLM must be interpreted in light of the information asymmetry: RF trains on~960 labeled examples from the target population per fold, while LLMs operate with zero training examples. That the zero-shot LLM closes to within 6 pp of a trained RF—achieving roughly 90% of the supervised ceiling—is arguably the more noteworthy finding. Model size does not predict accuracy: Qwen3.5 (27B) outperforms gpt-oss (120B) by 0.8 pp, while Gemma3 (also 27B) trails by 4.5 pp. Qwen3.5's advantage likely reflects its stronger Chinese-language training data, given that TEDS is conducted in Mandarin.

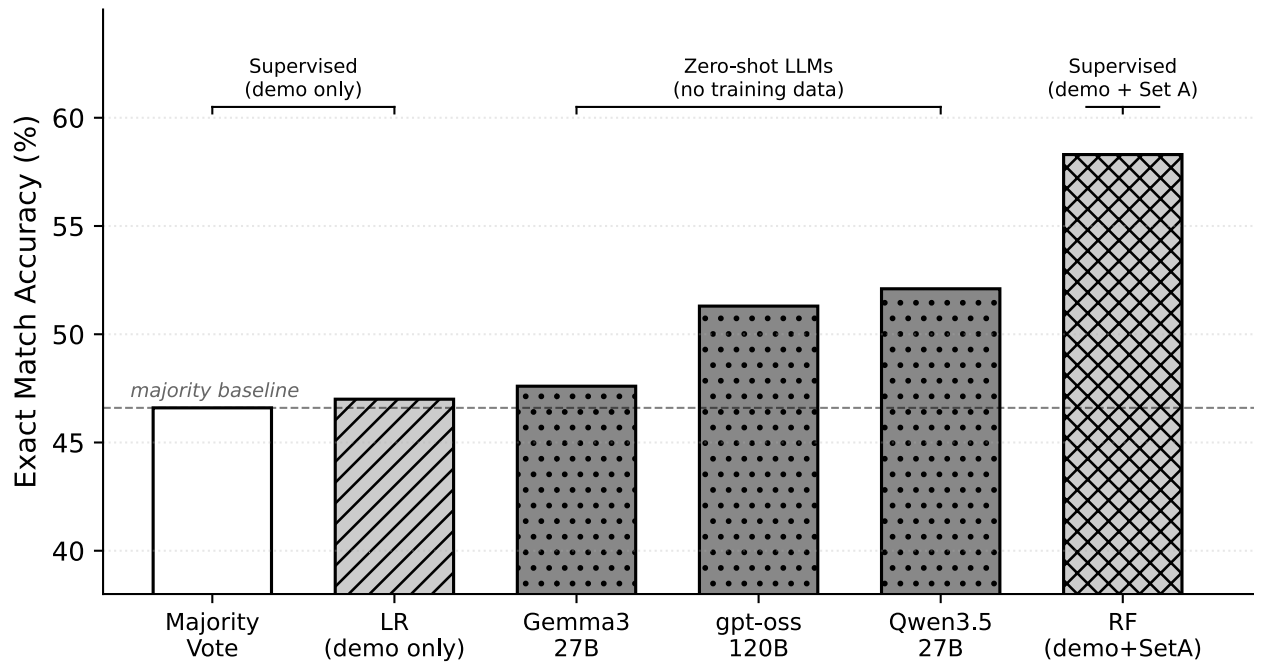

Fig. 1. Exact match accuracy across information regimes. Supervised ML baselines (hatched) train on same-population data via 5-fold CV; LLMs (dotted) operate zero-shot. The 6 pp gap between the best LLM and RF reflects the value of ~960 labeled training examples.

TABLE III. EXACT MATCH BY CONSTRUCT (LLM AVG). RF INCLUDED FOR CONPARISON

| | Construct | LLM | RF | Tier |
|---|---|---|---|---|
| 1 | Party competence | 67.2% | 66.3% | High |
| 2 | Gov't performance | 62.7% | 65.8% | High |
| 3 | Cross-strait relations | 58.6% | 70.5% | High |
| 4 | Political efficacy | 60.7% | 65.6% | High |
| 5 | Institutional trust | 60.4% | 64.4% | High |
| 6 | Gov't trust | 58.4% | 63.3% | Med. |
| 7 | Political system | 57.8% | 64.3% | Med. |
| 8 | Political participation | 51.5% | 59.5% | Med. |
| 9 | Democratic attitudes | 46.6% | 50.6% | Med. |
| 10 | Economic evaluation | 44.4% | 51.6% | Med. |
| 11 | Party favorability | 29.5% | 40.8% | Low |
| 12 | Candidate evaluation | 24.7% | 33.4% | Low |
| 13 | Sovereignty | 23.3% | 56.1% | Low |

### B. Construct-Level Predictability (RQ2)

Table III reveals a consistent hierarchy across all models. We interpret this through the lens of Converse's [16] attitude constraint framework: constructs tightly linked to partisan structure exhibit higher predictability, while constructs involving personal judgment show weaker constraint and lower accuracy.

The top tier (LLM >60%) comprises constructs tightly coupled to party identification: party competence, government performance, and institutional trust. Party ID—available as a demographic variable—effectively determines these positions. The hierarchy is remarkably consistent across models (cross-model range typically <8 pp), suggesting it reflects genuine attitudinal structure rather than model-specific artifacts.

The middle tier (40–60%) includes constructs requiring more nuanced reasoning. Political participation (B3) involves context-dependent engagement levels. Economic outlook (F2) requires integrating retrospective evaluations with forward-looking projections. Democratic attitudes (D15) involve moral reasoning about authoritarian leadership, weakly constrained by partisanship.

TABLE IV. VARIANCE RATIO BY ITEM FOR LLMS (GPT=GPT-OSS, QW=QWEN3.5, GE=GEMMA3) AND RF. VALUE < 1 = COLLAPES.

| Item | True σ | gpt | Qw | Ge | RF |
|---|---|---|---|---|---|
| D15 (strong Idr.) | 1.05 | 0.32 | 0.44 | 0.46 | 0.20 |
| D11e (trust pty.) | 0.73 | 0.44 | 0.60 | 0.42 | 0.66 |
| D5 (tax waste) | 0.74 | 0.73 | 0.62 | 0.43 | 0.72 |
| C2 (Tsai pandemic) | 0.93 | 1.01 | 0.95 | 0.87 | 0.92 |
| P2b (DPP fav.) | 2.82 | 0.96 | 1.09 | 0.83 | 1.00 |
| M5a (sovereignty) | 2.36 | 1.24 | 1.12 | 0.82 | 0.69 |
| **Average** | – | **0.85** | **0.84** | **0.67** | **0.72** |

The bottom tier (<40%) includes 0–10 scale items where exact match is arithmetically harder, but the sovereignty item (M5a) deserves special attention: LLMs achieve only 23.3% while RF reaches 56.1%—the largest LLM–ML gap (33 pp). This independence–unification scale is the most politically sensitive item in Taiwanese politics and likely triggers safety-related hedging in LLMs. RF, lacking such constraints, exploits statistical patterns directly.

### C. Variance Collapse and Safety Alignment (RQ3)

Two patterns observed in the silicon sampling literature are the systematic narrowing of LLM response distributions relative to human ones [4]—which we term variance collapse—and behavioral effects of safety alignment [13]. Our results complicate both narratives.

All methods exhibit variance collapse (Table IV), but the severity differs in unexpected ways. LLMs (VR=0.67–0.85) are comparable to or better than RF (VR=0.72). The shared collapse across LLMs and supervised models is consistent with a partly structural property of the prediction task—predicting individual survey responses from limited context inherently biases toward population means—rather than solely an LLM artifact. Counterintuitively, LLMs produce more realistic distributional diversity despite lower point accuracy—a tradeoff relevant for survey augmentation, where population-level heterogeneity may matter more than individual-level prediction. Collapse severity indicates the construct hierarchy: partisan items maintain near-normal variance, while personal judgment items collapse severely across all methods.

Table V compares aligned and abliterated model variants. The effects are surprisingly model-dependent: gpt-oss loses 6.1 pp accuracy after abliteration, Gemma3 loses 2.3 pp, and Qwen3.5 is essentially unchanged (−0.1 pp). Abliteration also affects variance in inconsistent directions: gpt-oss's VR increases (0.85→0.90), suggesting alignment was con straining its output diversity, while Gemma3's VR decreases (0.67→ 0.45), indicating alignment was actually helping maintain diversity. This heterogeneity challenges the assumption that safety alignment has uniform effects on output distributions [13], [14].

TABLE V. ALIGNED VS. ABLITERATED (ALIGNMENT-REMOVED) MODEL VARIANTS.

| | Exact Match | | | Variance | |
|---|---|---|---|---|---|
| **Model** | **Align.** | **Ablit.** | **Δ** | **Align.** | **Ablit.** |
| gpt-oss 120B | 51.3% | 45.2% | -6.1 | 0.85 | 0.90 |
| Qwen3.5 27B | 52.1% | 52.0% | -0.1 | 0.84 | 0.76 |
| Gemma3 27B | 47.6% | 45.4% | -2.3 | 0.67 | 0.45 |

At the item level, the sovereignty item M5a is instructive: gpt-oss aligned achieves 28.2% but drops to 18.6% after abliteration, while Gemma3 improves from 12.0% to 18.5%. This suggests gpt-oss's alignment provided useful calibration for sensitive political items, while Gemma3's alignment may have actively distorted its predictions. Researchers should verify alignment effects per model rather than assuming universality.

## VI. DISCUSSION

Our results paint a nuanced picture of silicon sampling's capabilities and limitations, with implications for both methodology and practice.

**LLMs as zero-shot survey simulators.** Zero-shot LLMs achieve 52% exact match on genuinely unseen items—meaningful given the task's difficulty, but 6 pp below the RF model using same-population training data. On a task where random guessing yields ≈25% and majority vote yields 47%, an LLM with no training examples from this population closes much of the gap toward a supervised model's 58%. The practical implication is that LLMs are not yet substitutes for statistical models when labeled data exists, but they offer unique zero-shot applicability to new populations without retraining.

**A theory-grounded predictability hierarchy.** The construct-level results are theoretically interpretable through Converse's [16] attitude constraint framework. Partisan-constrained attitudes are highly predictable because party identification structures belief systems. Weakly constrained attitudes require idiosyncratic information unavailable in 28 context items. This hierarchy is consistent across all three LLM families and the RF baseline, suggesting it reflects genuine attitudinal structure.

**The sovereignty anomaly.** The sovereignty item M5a shows the sharpest divergence: RF reaches 56% accuracy, while LLMs average only 23%. We interpret this gap primarily as evidence that the 11-point independence–unification scale is structurally difficult to predict in a cross-survey setting, likely because it captures a multidimensional stance rather than a simple binary continuum. Safety hedging may further depress LLM performance, but this explanation remains secondary and tentative given the mixed abliteration evidence across model families.

**Rethinking variance collapse.** Counterintuitively, LLMs better preserve distributional diversity (VR=0.84–0.85) than RF (VR=0.72), challenging the narrative that variance collapse is primarily an LLM artifact [4]. For survey augmentation applications where population-level heterogeneity matters more than individual accuracy, LLMs may actually be preferable.

**Non-WEIRD generalization.** Our Mandarin-language results (best LLM: 52.1%) are broadly comparable to English-language findings [2], and Qwen3.5's advantage over Gemma3 suggests that language-specific training data improves survey simulation—relevant for researchers in non-English contexts [6], [7].

**Practical implications.** For exploratory purposes—piloting survey instruments or generating hypotheses—cross-survey transfer provides a rigorous diagnostic. For production use as a survey replacement, the current accuracy levels are insufficient. The most promising near-term application may be hybrid approaches combining LLM-generated priors with smaller human samples [15].

## VII. Limitations and Future Work

Several limitations warrant discussion. First, the Set A/B partition is fixed and theory-driven rather than optimized; ablation studies on Set A size would quantify the minimum context needed for reliable transfer. Second, we evaluate only open-weight models; closed-source models (GPT-4, Claude) may perform differently but preclude abliteration analysis—a tradeoff we consider worthwhile for reproducibility. Third, items on 0–10 scales are arithmetically disadvantaged in exact-match evaluation relative to 4-point Likert items, though our within-±1 and VR metrics partially address this. Fourth, we lack human test-retest reliability estimates for TEDS, which would provide a natural accuracy ceiling. Fifth, the 5 pp improvement over majority baseline, while statistically significant across >8,000 item-level predictions, represents modest practical utility for individual-level prediction.

Future work should explore calibrated decoding, fine-tuning on target-population data [15], richer prompting strategies such as interview transcripts [8], and cross-cultural replication in other non-WEIRD contexts. Multi-model ensembles combining LLM predictions with supervised baselines may offer the best of both worlds.

## VIII. Conclusion

We proposed cross-survey transfer as a rigorous evaluation framework for silicon sampling and tested it with three LLMs, abliterated variants, and supervised baselines on Taiwanese political survey data. Three key findings emerge: (1) zero-shot LLMs achieve meaningful but modest accuracy (52%) on genuinely unseen items, trailing supervised models (58%) that benefit from same-population training; (2) a stable construct predictability hierarchy—from partisan attitudes (67%) to personal judgment (23%)—reflects genuine attitudinal structure; and (3) two commonly cited limitations—variance collapse and alignment distortion—are more nuanced than previously reported: collapse affects all prediction methods, LLMs actually better preserve distributional diversity, and alignment effects are model-dependent rather than universal. These findings position cross-survey transfer as a diagnostic tool for understanding when silicon sampling works—and when it does not.

## Acknowledgment

The authors thank the Election Study Center at National Chengchi University for making the TEDS 2024 dataset publicly available. Code and analysis scripts are available upon request.